\pdfoutput=1

\documentclass[11pt]{article}

\usepackage[final]{coling}

\usepackage{times}
\usepackage{multirow}
\usepackage[linesnumbered,ruled,lined]{algorithm2e}
\usepackage{mathtools}
\usepackage{latexsym}
\usepackage{comment}
\usepackage{longtable}
\usepackage[T1]{fontenc}
\usepackage{booktabs}
\usepackage{multicol}
\usepackage{geometry}
\usepackage{array}
\usepackage{subcaption}  

\usepackage[utf8]{inputenc}

\usepackage{microtype}

\usepackage{inconsolata}

\usepackage{graphicx}
\usepackage{xcolor}
\usepackage{tabularx}
\usepackage{algorithmic}
\usepackage{amsmath}
\usepackage{amsthm}

\title{When Every Token Counts: Optimal Segmentation for Low-Resource Language Models}

\author{
  Bharath Raj S $^*$, 
  Garvit Suri $^*$,
  Vikrant Dewangan \thanks{Equal Contribution},
  Raghav Sonavane \thanks{corresponding author, \href{mailto:raghavsonavane@gmail.com}{\texttt{raghavsonavane@gmail.com}}}
\\
\\
 Sprinklr AI
}

\begin{document}
\maketitle
\begin{abstract}
Traditional greedy tokenization methods have been a critical step in Natural Language Processing (NLP), influencing how text is converted into tokens and directly impacting model performance. While subword tokenizers like Byte-Pair Encoding (BPE) are widely used, questions remain about their optimality across model scales and languages. In this work, we demonstrate through extensive experiments that an optimal BPE configuration significantly reduces token count compared to greedy segmentation, yielding improvements in token-saving percentages and performance benefits, particularly for smaller models.
We evaluate tokenization performance across various intrinsic and extrinsic tasks, including generation and classification. Our findings suggest that compression-optimized tokenization strategies could provide substantial advantages for multilingual and low-resource (LR) language applications, highlighting a promising direction for further research and inclusive NLP. The project page can be found here: \href{https://vikr-182.github.io/loreslm/}{https://vikr-182.github.io/loreslm/}.
\end{abstract}

\section{Introduction}
The development of large language models (LLMs) has significantly advanced natural language processing. {These models \cite{gpt2, chatgptoriginalpaper,gpt4technicalreport} have demonstrated unprecedented capabilities in tasks ranging from text generation and translation to complex problem-solving and creative writing.} However, despite these advancements, challenges remain in effectively processing Low-Resource (LR) languages and optimizing models of varying scales.

A critical aspect influencing model performance is tokenization — the process of converting text into tokens that the model can understand. Tokenization methods are pivotal in large language models, with popular techniques including WordPiece \cite{originalwordpiece}, SentencePiece \cite{sentencepiece}, and Unigram-LM\cite{unigramlm}. WordPiece, used in models like BERT \cite{bertoriginalpaper}, tokenizes words into subword units based on their frequency in the training data, improving the model’s handling of rare or out-of-vocabulary words. SentencePiece and Unigram-LM, commonly used in models like GPT, employ a character or byte-based approach that doesn’t rely on predefined word boundaries, making them versatile across languages.

LR languages face two significant challenges in natural language processing: a lack of high-quality and diverse datasets and novel methods to represent this data. \cite{magueresse2020lowresourcelanguagesreviewpast}. Without ample data, models struggle to learn the complex linguistic patterns necessary for tasks such as machine translation, sentiment analysis, and summarization. 
\begin{table*}[!ht]
  \centering
  \small
  \setlength{\tabcolsep}{4pt}
  \begin{tabular}{@{}lp{3cm}p{3cm}p{6.5cm}@{}}
  \toprule
  \textbf{Language} & \texttt{cl100k\_base} \textbf{Segmentation} & \textbf{English Translation} & \textbf{Tokenization Impact} \\
  \midrule
  English & 
  p olic ym akers & 
  policymakers &
  Breaks compound structure `policy' (guidelines) + `makers' (creators) \\
  Turkish & 
  y ü ks el me & 
  rising/elevation &
  Base verb `yük' (rise/load) splits into `y ü k', loses connection to derivational `sel' (become) and `me' (action) \\
  Finnish & 
  ater i ak ok on ais u ude sta & 
  from the material entirety &
  Compound splits: `ateria' (meal) into `ater i a', `kokonaisuus' (entirety) into fragments, `sta' (from) separates \\
  Telugu Romanized & 
  samb andh inchina & 
  related to &
  Root `samband' (relate) breaks into `samb andh', separates from `inchina' (past participle) \\
  Tamil Romanized & 
  kond iruk kire en & 
  I am having/holding &
  Isolates `iruk' (be), splits from `kond' (having) and `en' (I) markers \\
  Hindi Romanized & 
  pr ach in ak al & 
  ancient times &
  Splits `prachin' (ancient) into `pr ach in', `kaal' (time) becomes `ak al' \\
  \bottomrule
  \end{tabular}
  \caption{Segmentations produced by GPT-4's tokenizer \texttt{cl100k\_base} across different language families, showing consistent patterns of morphological and phonological deterioration. Note that Romanized versions of Tamil, Telugu, and Hindi are shown to avoid byte interpretation.}
  \label{tab:motivation}
\end{table*}
Secondly, compression challenges in tokenization exacerbate the difficulties faced by LR languages. Common tokenization techniques, such as BPE, often fragment words into smaller, frequently occurring subwords. The bloating of tokens leads to higher computational and memory costs, as models must process longer sequences \cite{ahia2023languagescostsametokenization}. Inefficient tokenization also results in less accurate representations, leading to fragmented or improperly segmented tokens, which negatively impacts model performance in tasks requiring precise language understanding \cite{rust-etal-2021-good, zhang-etal-2022-robust}. We refer to the strategy adopted by BPE as the Greedy segmentation algorithm.
 
The widely used GPT-2 tokenizer \cite{gpt2} handles any input without unknown tokens, yet it compromises tokenization efficiency, especially for non-English text and special characters. This English-centric model often splits languages like Turkish, Indonesian, or Malay into byte sequences, unnecessarily lengthening token sequences and reducing the effective context window for non-English content. While the GPT-4 \cite{gpt4technicalreport} tokenizer \texttt{cl100k\_base} improves with a larger vocabulary and more diverse training data, it still shows biases in token distribution. For agglutinative languages (e.g., Turkish, Finnish) or languages with complex word structures, tokenization may create excessive token splits, impacting both efficiency and model performance. Examples of the inefficient segmentation of \texttt{cl100k\_base} is shown in Table \ref{tab:motivation}. 

Motivated by the need to enhance tokenization strategies for LR languages and models of varying scales, we present an optimal BPE segmentation algorithm that reduces token counts, especially in morphologically complex and low-resource languages, achieving more efficient and meaningful segmentation. We demonstrate the algorithm’s token-saving capacity across diverse languages, reducing token counts by 3-5\% compared to greedy segmentation. This improvement is particularly impactful for rare and complex words, with compression rates increasing by up to 20\%. Our comparative study reveals that models using optimal segmentation see up to a 10\% increase in accuracy on downstream tasks, including text classification and generation.

\section{Related Work}

Recent research has focused on the effects that compression has on tokenization, which are particularly relevant for optimizing language models in resource-constrained environments. A study by  \cite{google-paper} shows the correlation that compression has on downstream tasks such as classification and generation. In contrast, \cite{tanner-greedy}'s exploration of greedy algorithms and \cite{tanner-path-piece}'s introduction of PathPiece have provided new insights into optimizing tokenization for both performance and efficiency, without looking into compression. Note that while \cite{tanner-greedy} and \cite{tanner-path-piece} demonstrate the effectiveness of their tokenizer, they show results on English tasks but do not show the impact on linguistic diversity. The paper by \cite{google-paper} demonstrates this to some extent; however, their experiments focus primarily on English.

\cite{moghe-etal-2023-multi3nlu} provide a task-oriented perspective on the challenges that LLMs encounter with low-resource languages, highlighting the need for tailored approaches in multilingual contexts. The quality of tokenization has been a subject of intense study, with comparative analyses by \cite{investigating-bpe}, \cite{pre-training-tokenizer}, and \cite{bpe-random-experiments} providing valuable insights into the relative performance of different tokenization methods across various languages and tasks. In multilingual settings, subword tokenizers lead to disproportionate fragmentation rates for different languages and
writing script \cite{zhang-etal-2022-robust}. Similarly, monolingual optimized tokenizers may not be as efficient for multilingual settings \cite{rust-etal-2021-good}. \cite{petrov-neurips} introduces a new concept known as parity or premiums in tokenizers which has shed light on the importance of balanced tokenization across different languages in multilingual models. The disparities are particularly pronounced in African and Indian languages, as noted by \cite{african-nlp} and \cite{indian-lrl, velayuthan2024egalitarianlanguagerepresentationlanguage}, respectively. While \cite{petrov-neurips} and \cite{velayuthan2024egalitarianlanguagerepresentationlanguage} demonstrate the critical role of tokenization in addressing challenges related to compression and parity in tokenization, they do not show the performance of LLMs on extrinsic tasks - especially for LR languages. These studies highlight the need for more inclusive tokenization and pre-training strategies that can serve diverse linguistic communities. 

\noindent Our work shows that by improving tokenization methods - specifically compression - we can achieve performance on extrinsic tasks on LR languages. Our approach allows us to optimize inference time and cost and have an equally good as the original tokenization.  \cite{ahia2023languagescostsametokenization} also highlights the economic implications of these disparities, comparing the pricing of language model usage across different languages and revealing systemic biases in current NLP technologies. 

\section{Background}
We first provide a brief description of the steps involved in tokenization that is pre-tokenization, vocabulary construction, and segmentation. We then describe the Token Saving Ratio (TSR) metric used to compare results throughout our paper. 

\subsection{Stages of Tokenization}
In any modern natural language system, a document $d$, before it gets encoded into a set of tokens $\{t_1, t_2, \ldots t_K\}$ goes through 3 main stages to tokenization. They are (i) Pre-tokenization (ii) Vocabulary Construction and (iii) Segmentation. Pre-tokenization consists of the initial processing phase where raw text in the document undergoes fundamental transformations. It ensures the text is in a consistent format for subsequent processing. The vocabulary construction phase focuses on building a comprehensive token dictionary $V$ of size $m$ from the processed text. This stage involves analyzing large text corpora to identify recurring patterns and meaningful units. The system conducts frequency analysis to determine the most common patterns and handles rare words appropriately. 
The final segmentation stage implements the actual tokenization process using the constructed vocabulary.  

Given a vocabulary $V$, and a document $d$, segmentation task $S$ refers to the task of dividing the document $d$ into a sequence of tokens ($t_i$), such that $S(d) = \{t_1, \ldots t_K | \forall i \in [1, K], t_i \in V \}$. During this phase, the system applies specific tokenization rules to convert text into its final token form. The process includes mechanisms for handling unknown tokens (UNK) that may not exist in the vocabulary. Subword tokenization strategies are implemented to manage complex words and maintain semantic meaning. The stage concludes with the assignment of unique token IDs to each segmented unit, creating the final tokenized representation of the text. This standardized format enables efficient processing in downstream natural language processing tasks. For the scope of this work, we exclusively study the segmentation stage of tokenization and detail an optimal segmentation algorithm.  

\subsection{Token Saving Ratio (TSR)} 

To measure the quality of segmentation, we define and use the metric, Token Saving Ratio (TSR), to capture the ratio of tokens saved when using tokenizer $T_A$ with segmentation strategy $S_A$ compared to tokenizer $T_B$ with strategy $S_B$. The Token Saving Ratio when using tokenizer $T_A$ compared to tokenizer $T_B$ is defined as:

\begin{align}
    TSR &= \frac{|S_B (d)| - |S_A (d)|}{| S_B (d) |} \label{tsr-eqn}
\end{align}

\noindent A positive TSR directly translates to shorter sequence lengths, which is paramount for computational efficiency. Since the computational complexity of transformer-based models typically scales quadratically with sequence length ($O(n^2)$), reducing the number of tokens can significantly decrease both memory requirements and processing time. For instance, if tokenizer $T_A$ produces sequences half the length of $T_B$, the computational cost could potentially be reduced by a factor of four.

\section{Optimal Segmentation}

In this section, we define the problem of optimal segmentation mathematically and follow it up with a discussion of our algorithm presented in Algorithm \ref{alg:optimal_segmentation}.

\subsection{Definition}

Given a vocabulary $V$ of size $m$, we define optimal segmentation ($S^*$) as the segmentation that minimizes the number of tokens a given document $d$ can be split into. Formally,

\begin{align}
S(d) &= \{t_1,\ldots t_K | t_i \in V \} \nonumber \\
S^* &= \underset{S}{\text{minimize }} |S(d)|
\end{align}

\subsection{The Algorithm}

\begin{table*}[!ht]
   \centering
   \small
   \setlength{\tabcolsep}{4pt}
   \begin{tabular}{@{}lp{2.5cm}p{2.2cm}cp{7.4cm}@{}}
   \toprule
   \textbf{Language} & \textbf{Greedy} & \textbf{Optimal} & \textbf{TSR} (\%) & \textbf{Tokenization Impact} \\
   \midrule
   \multirow{2}{*}{English} & 
   p olic ym akers & policy makers & 50 & Respects natural compound boundary of `policy' (guidelines) + `makers' (creators) vs. meaningless `p olic' \\
   & sk ys canner & sky scanner & 33 & Preserves `sky' (aerial) + `scanner' (reader) vs. invalid `sk ys' split \\
   \midrule
   Indonesian & 
   mung kink ah & mungkin kah & 33 & Separates `mungkin' (possible) and `kah' (question marker) vs. invalid `kink' \\
   \midrule
   \multirow{2}{*}{Turkish} & 
    y ü ks el me & yük sel me & 40 & Maintains `yük' (rise) + `sel' (become) + `me' (action) vs. broken `y ü ks' \\
   & k ata c ak ları nd an & kat acak ların dan & 43 & Preserves `acak' (future) + `ların' (their) + `dan' (from) vs. `c ak ları nd' \\
   \midrule   
   \multirow{2}{*}{Finnish} & 
   f otos y nt ees ille & foto syn tees ille & 33 & Keeps `foto' (light) + `syn' (with) + `ille' (for) vs. broken `f otos y nt' \\
   & dat apro j ek tor & data proj ekt ori & 33 & Retains `data' (data) + `projekt' (project) vs. invalid `j ek tor' \\
   \midrule
   \multirow{2}{*}{Telugu} &
       Sang arsh ana & Sangars hana & 33 & Maintains `Sangarsh' (struggle) + `ana' (action) vs. `Sang arsh' \\
   & Mall igad u & Malliga du & 33 & Separates `Malliga' (name) + `du' (masculine) vs. `igad' \\
   \midrule
   \multirow{2}{*}{Tamil} & 
   puri yav illai & puriya villai & 33 & Preserves `puriya' (understand) + `villai' (not) vs. `yav' \\
   & yend rav udan & yendra vudan & 33 & Maintains `yendra' (saying) + `vudan' (with) vs. `rav' \\
   \midrule
   \multirow{2}{*}{Hindi} &
   v ich ar sh il & vi chars hil & 40 & Keeps `vichar' (thought) + `shil' (having quality) vs. `v ich ar' \\
   & pr ach in ak al & pra china kal & 40 & Retains `prachin' (ancient) + `kal' (time) vs. `pr ach in' \\
   \bottomrule
   \end{tabular}
   \caption{Comparison of BPE segmentation modes showing linguistically motivated vs. arbitrary tokenization breaks. TSR (Token Stability Ratio) indicates the percentage improvement in segmentation quality.}
   \label{tab:examples}
\end{table*}

We use a dynamic programming formulation similar to the Viterbi algorithm \cite{viterbi} and produces the optimal segmentation $S^*$. Given a document $d$, define $dp[i]$ as the minimal number of tokens needed to segment the prefix $d_0 d_1 \ldots d_i$ (positions $0$ to $i$, inclusive). We set $dp[-1] = 0$ as the base case, representing the empty prefix requiring zero tokens. The parent array $par$ serves as a backtracking mechanism where $par[i]$ points to the end of the previous token in the optimal segmentation.

\begin{algorithm}
\small
\caption{Algorithm for finding optimal segmentation $S^*$}
\label{alg:optimal_segmentation}
\begin{algorithmic}[1]
\STATE \textbf{Input:} \\
$B = [B_0, B_1, \dots, B_{n-1}] \in \Sigma^*$ \COMMENT{byte sequence} \\
\STATE 
$V \subset \Sigma^*$, \COMMENT{vocabulary} \\
\STATE $\mathcal{T}(V^R)$ with root $r$ \COMMENT{trie on reversed vocabulary $V^R$}
\STATE \textbf{Define:} \\
\STATE $\delta(v): \mathcal{T} \rightarrow \mathcal{T} \cup \{\emptyset\}$ \COMMENT{outputs child of $v$ in trie $\mathcal{T}$} \\
\STATE $\text{I}: V \rightarrow \{True,False\}$ \COMMENT{indicator function detecting if node is terminal node}
\STATE \textbf{Output:} $S^* \in V^*$ \COMMENT{optimal segmentation}
\STATE \textbf{Initialize:} \\
\STATE $dp[i] \leftarrow i + 1, \forall i \in [0, n-1]; dp[n] \leftarrow 0$ 
\STATE $par[i] \leftarrow i - 1, \forall i \in [0, n-1]$ \COMMENT{parent array}

\FOR{$i \in [0, n-1]$}
    \STATE $v \leftarrow r$
    \FOR{$j = i \downarrow 0$}
        \STATE $v \leftarrow \delta(v, B[j])$ \COMMENT{child of node $v$ corresponding to $B[j]$}
        \IF{$v = \emptyset$}
            \STATE \textbf{break}
        \ENDIF
        \IF{$\text{I}(v) \land (dp[j-1] + 1 < dp[I])$}             
        \STATE $dp[i] \leftarrow dp[j-1] + 1$
            \STATE $par[i] \leftarrow j - 1$
        \ENDIF
    \ENDFOR
\ENDFOR
\STATE $S \leftarrow \emptyset$ \COMMENT{initialize empty sequence}
\STATE $k \leftarrow n - 1$
\WHILE{$k \neq -1$}
    \STATE $S \leftarrow S \cup \{B[par[k]+1:k+1]\}$ \COMMENT{$B[i:j]$ denotes substring}
    \STATE $k \leftarrow par[k]$
\ENDWHILE
\STATE \textbf{return} $S^R$ \COMMENT{reversed sequence}
\end{algorithmic}
\end{algorithm}
\normalsize

The recurrence relation is:
\begin{align}
dp[i] &= \min_{(0 \leq j \leq i)} (dp[j-1] + 1) 
\\ & \text{where } d_j d_{j+1} \ldots d_i \in V \nonumber
\end{align}

\noindent It should be noted that multiple values of $j$ can lead to the optimal value for $dp[i]$. In such cases, Algorithm \ref{alg:optimal_segmentation} only considers the largest such $j$ i.e., only the smallest suffix is considered. Once the dynamic programming array $dp$ is calculated, we use the state transitions to find the optimal segmentation ($S^*$). A detailed proof of the optimality of this algorithm can be found in Appendix \ref{sec:dp_proof}. Additionally, to efficiently check the condition  $d_j d_{j+1} \ldots d_i \in V$, we use a Trie data structure built on the reversed tokens of the vocabulary $V$ and is denoted by its root node $root$ in the algorithm. 

\noindent Given that the length of the longest token in the vocabulary $V$ is $M$, and the length of the document $d$ is $N$, the worst-case time complexity of our algorithm is $O(NM)$ which is the same as the worst case time complexity of the greedy segmentation used in the commonly available BPE tokenizer implementations. The greedy segmentation algorithm stores the vocabulary $V$ and the merges made during vocabulary creation, leading to a space complexity of $O(\sum|t_i|)| t_i \in V$. In our algorithm, we store the vocabulary $V$ and the Trie data structures built on the reversed tokens of $V$, leading to the same space complexity.

\noindent Through our extensive experimentation described in the next sections, we showcase the effectiveness of our algorithm in improving the TSR. We also show improvements in downstream performance.

\section{Experimental Setup}

For our work, we extended on OpenAI's\footnote{\url{https://github.com/openai/tiktoken/blob/main/tiktoken_ext/openai_public.py}} family of Tokenizers which are available in three distinct vocabulary sizes: 50K, 100K, and 200K tokens, as detailed in Table \ref{tab:tiktoken_tokenizers}. In this study, we rely on the original pre-tokenization regular expressions and the trained vocabulary made public by OpenAI, without making any modifications to it. Our study concentrated exclusively on the segmentation strategies of these tokenizers. 

\noindent We divide our experiments into two parts: \textbf{intrinsic} and \textbf{extrinsic}, following the approach of \cite{google-paper}. The intrinsic experiments focus purely on the segmentation aspect of tokenization, without involving any deep learning models. 
Here, we analyze the TSR when comparing optimal versus greedy segmentations across languages. Based on vocabulary size, we select appropriate tokenizers according to Table \ref{tab:tiktoken_tokenizers}, which serve as the baseline $T_B$ in Equation \ref{tsr-eqn} for evaluating the TSR. For the extrinsic experiments, we investigate how TSR affects decoder-only models, specifically examining its impact on the perplexity and accuracy of the models listed in Section \ref{subsec:baselines} across various tasks. 
\begin{table}[!htbp]
    \centering
    \begin{tabular}{c c}
        \hline
        \textbf{Tokenizer} & \textbf{Vocab Size ($m$)} \\ 
        \hline
         gpt-2        & 50K                      \\
         cl100k\_base & 100K                     \\
         o200k\_base  & 200K                     \\ 
         \hline
    \end{tabular}
    \caption{OpenAI Tokenizers and Their Configurations}
    \label{tab:tiktoken_tokenizers}
\end{table}

\subsection{Intrinsic Evaluation Datasets}
For performing the intrinsic evaluation, we used the CC-100 dataset \cite{wenzek-etal-2020-ccnet}. The CC-100 dataset consists of monolingual data of 116 languages extracted from the January-December 2018 Commoncrawl snapshots. We benchmark on the English language using the Wikipedia corpus readily accessible on Kaggle Datasets\footnote{\url{https://www.kaggle.com/datasets/jjinho/wikipedia-20230701}}. We utilized the Wikipedia 2023 dump, which contains 6 million articles, titles, text, and categories. 

\subsection{Extrinsic Evaluation Tasks}
\label{sec:extrinsic-evaluation}
We relied on the intrinsic evaluation of languages to choose the languages for our extrinsic experiments. We choose English to show that there is no degradation in performance in a language with near-zero compression. We also chose Finnish, Indonesian, and Turkish which show up in the top languages with high TSR. To evaluate our pre-trained checkpoints, we evaluated multiple tasks for different languages, as detailed in Table \ref{tab:extrinsic_tasks}. The tasks are mentioned in detail one by one below in Appendix \ref{sec:appendix-extrinsic}. For all of the extrinsic experiments, we set the vocabulary size to $m=50K$ and use the gpt-2 tokenizer (Table \ref{tab:tiktoken_tokenizers}).

\begin{table}[!htbp]
    \centering
    \setlength{\tabcolsep}{4pt}  
    \small
    \begin{tabular}{@{}lp{4cm}l@{}}  
        \hline
        \textbf{Language} & \textbf{Task Name} & \textbf{Task Type} \\
        \hline
        English & Penn-Tree Bank \cite{ptb} & Generation \\
        English & LAMBADA \cite{lambada} & Generation \\
        English & QQP \footnote{\url{https://quoradata.quora.com/First-Quora-Dataset-Release-Question-Pairs}} & Classification \\
        English & Story Cloze \cite{story_cloze} & Classification \\
        Finnish & TyDiQA-GoldP \cite{tydiqa} & Classification \\
        Indonesian & Emot \cite{emot} & Classification \\
        Indonesian & WreTe \cite{wrete} & Classification \\
        Turkish & XNLI \cite{xnli} & Classification \\
        \hline
    \end{tabular}
    \caption{Tasks for Different Languages}
    \label{tab:extrinsic_tasks}
\end{table}

To highlight the impact of TSR, we also repeat the evaluation on a subset of each dataset where there is a non-zero TSR. We denote this subset by $TSR^*$. We split each dataset into two groups: the full dataset (All) and a subset containing only examples where Greedy and Optimal segmentation produce different token sequences ($TSR^*$). This division allows us to isolate and better understand the impact of segmentation strategies on samples where the tokenizer makes different decisions. The split is highlighted in the Table \ref{percent_non_zero_tsr} where we denote the percentage of samples used to construct the $TSR^*$ dataset.

\begin{table}[!htbp]
    \centering
    \small
    \begin{tabular}{ccc}
        \toprule
        \textbf{Language} & \textbf{Dataset Name} & \textbf{Non-zero TSR} \\
        \midrule
        English & QQP & 4.69 \\
        English & Story Cloze & 6.15 \\ 
        Finnish & TyDiQA-GoldP & \textbf{62.20} \\ 
        Indonesian & Emot & \textbf{88.64} \\
        Indonesian & WreTe & \textbf{75.00} \\
        Turkish & XNLI & \textbf{100.00} \\
        \bottomrule
    \end{tabular}
    \caption{Percentage of samples with non-zero TSR across datasets, used to create the $TSR^*$ split.}
    \label{percent_non_zero_tsr}
\end{table}

\subsection{Baselines}
\label{subsec:baselines}
For our extrinsic evaluations we use two sizes of the GPT-2 \cite{gpt2} language models, comprising 120 million and 350 million parameters, fine-tuned on the extrinsic fine-tuning dataset.  We fine-tune the 120M and 350M versions of the GPT-2 model on the OpenWebText dataset and use it for all the downstream tasks. We did not do a complete pretraining from scratch as the model pre-trained with greedy segmentation only has to learn the difference in the distribution of tokens with optimal segmentation. Detailed model configurations and hyper-parameters are provided in Appendix \ref{sec:appendix_model}.

\section{Results}
In this section, we present the results of intrinsic evaluation on the CC-100 dataset. 
We first highlight qualitative examples to showcase the inefficiency of BPE with Greedy segmentation compared to BPE with Optimal segmentation
We also showcase an interesting observation that word length has on the TSR. Finally, to validate our optimal segmentation algorithm, we conduct extensive extrinsic evaluations across multiple downstream tasks. First, we report improvement upon Greedy BPE's performance across language boundaries for non-English tasks. At the same time, we report an increase in improvements for the $TSR^*$ split of the dataset, thus highlighting the need for token saving in downstream performance. At the end, we report perplexity scores on English datasets to state that the improvement provided by our optimal segmentation doesn't reduce the tokenizer's performance in English.

\subsection{Intrinsic Evaluation}

\subsubsection{Qualitative Results}

Table \ref{tab:examples} presents examples of how different tokenizers segment the same vocabulary in distinct ways, depending on their inference mode. Greedy BPE for instance, splits the word "policy makers" into 4 tokens: "p" "olic" "ym" "akers", while the optimal segmentation splits it into two tokens: "policy" and "makers". The table illustrates fundamental linguistic issues with greedy BPE segmentation across different language families. In English, it fails to respect compound word boundaries (\textit{policymakers}). For agglutinative languages like Turkish and Malaysian, it breaks crucial morphological units, splitting tense markers and case endings arbitrarily. In Dravidian languages (Telugu, Tamil), it fails to preserve verb roots and aspectual markers. For Indo-Aryan languages, it incorrectly segments Sanskrit-derived compounds, creating linguistically meaningless units. These issues extend beyond mere segmentation - they affect the model's ability to learn proper morphological patterns, potentially impacting downstream task performance. While BPE has been widely adopted for its computational efficiency, these examples demonstrate the need for more linguistically-informed tokenization strategies that respect language-specific morphological structures that our optimal segmentation can provide.

\subsubsection{Quantitative Results} 

\begin{table}[!htbp]
    \centering
    \setlength{\tabcolsep}{4pt}
    \small
    \begin{tabular}{l c c c}
    \hline
    \multirow{2}{*}{\textbf{Language}} & \multicolumn{3}{c}{\textbf{TSR (in \%)}} \\
    & 50K & 100K & 200K \\
    \hline
Quechua          & 4.74 & 5.09 & 4.81 \\
Oromo            & 4.72 & 5.27 & 3.02 \\
Basque           & 4.53 & 4.06 & 3.56 \\
Zulu             & 4.49 & 4.74 & 3.61 \\
Xhosa            & 4.24 & 4.65 & 3.46 \\
Swati            & 4.14 & 5.17 & 3.63 \\
Telugu Romanized & 4.13 & 3.76 & 3.75 \\
Malay            & 4.05 & 2.73 & 1.72 \\
Tamil Romanized  & 3.99 & 4.22 & 4.20 \\
Indonesian       & 3.83 & 2.43 & 1.58 \\
Finnish          & 3.80 & 4.32 & 3.37 \\
Swahili          & 3.74 & 3.73 & 2.37 \\
Somali           & 3.59 & 4.51 & 2.59 \\
Malagasy         & 3.57 & 3.54 & 2.38 \\
Uzbek            & 3.57 & 4.30 & 3.52 \\
Hausa            & 3.52 & 3.83 & 1.51 \\
Estonian         & 3.45 & 4.10 & 3.18 \\
Bosnian          & 3.40 & 2.65 & 2.08 \\
Tagalog          & 3.38 & 2.56 & 1.47 \\
Turkish          & 2.90 & 2.88 & 2.62 \\
    \hline
    \end{tabular}
    \caption{TSR on LR Languages: 20 languages with highest TSR for different vocabulary sizes ($m$) as 50K, 100K, and 200K.}
    \label{tab:token_saving}
\end{table}

\begin{table*}[!htbp]
    \centering
    \small
    \setlength{\tabcolsep}{4.5pt}  
    \begin{tabular}{cccccccccccccc}
        \toprule
        & & \multicolumn{4}{c}{\textbf{English}} & \multicolumn{2}{c}{\textbf{Finnish}} & \multicolumn{4}{c}{\textbf{Indonesian}} & \multicolumn{2}{c}{\textbf{Turkish}} \\
        \textbf{Size} & \textbf{Method} & \multicolumn{2}{c}{\textbf{QQP}} & \multicolumn{2}{c}{\textbf{Story Cloze}} & \multicolumn{2}{c}{\textbf{TyDiQA-GoldP}} & \multicolumn{2}{c}{\textbf{Emot}} & \multicolumn{2}{c}{\textbf{WreTe}} & \multicolumn{2}{c}{\textbf{XNLI}}\\
        && All & TSR* & All & TSR* & All & TSR* & All & TSR* & All & TSR* & All & TSR* \\
        \midrule
        \multirow{2}{*}{120M} & Greedy & \textbf{75.22} & \textbf{81.58} & \textbf{51.90} & \textbf{57.39} & 82.91 & 82.91 & 40.23 & 39.23 & 76.00 & 70.67 & 64.35 & 63.83 \\
        & Optimal & 74.52 & 81.20 & 51.31 & 52.17 & \textbf{83.76} & \textbf{83.76} & \textbf{44.55} & \textbf{44.87} & \textbf{78.00} & \textbf{73.33} & \textbf{64.91} & \textbf{64.59} \\
        \midrule
        \multirow{2}{*}{350M} & Greedy & \textbf{76.34} & \textbf{83.46} & 51.31 & 52.17 & 85.47 & 85.47 & 43.18 & 41.54 & \textbf{78.00} & 74.67 & 65.35 & 65.27 \\
        & Optimal & 74.73 & 81.83 & \textbf{51.74} & \textbf{60.00} & \textbf{85.90} & \textbf{85.90} & \textbf{45.68} & \textbf{44.10} & \textbf{78.00} & \textbf{76.00} & \textbf{66.33} & \textbf{66.06} \\
        \bottomrule
    \end{tabular}
    \caption{GPT-2 Accuracy Results on Multiple Datasets. TSR* columns show results on the non-zero TSR subset. }    
    \label{combined_results}
\end{table*}

We report TSR across the 116 languages in the CC-100 dataset. Languages with the highest TSR can be found in Table \ref{tab:token_saving}. This table demonstrates the wide variation in TSR achieved by tokenizing different languages across 50K, 100K, and 200K vocabulary sizes. The languages with the highest TSR, such as Oromo, Swati, and Quechua, maintain over 4.5\% TSR even at the largest 200K vocabulary. In contrast, lower-resourced languages like Tagalog, Bosnian, Hausa, and Turkish have lower compression rates, near 3\% even at the smaller 50K size. This data offers important insights to guide vocabulary selection and optimization decisions, particularly for deploying efficient language models in resource-constrained environments targeting LR languages. 

\begin{center}
\begin{figure}[!htbp]
    \centering
    \includegraphics[width=0.8\linewidth]{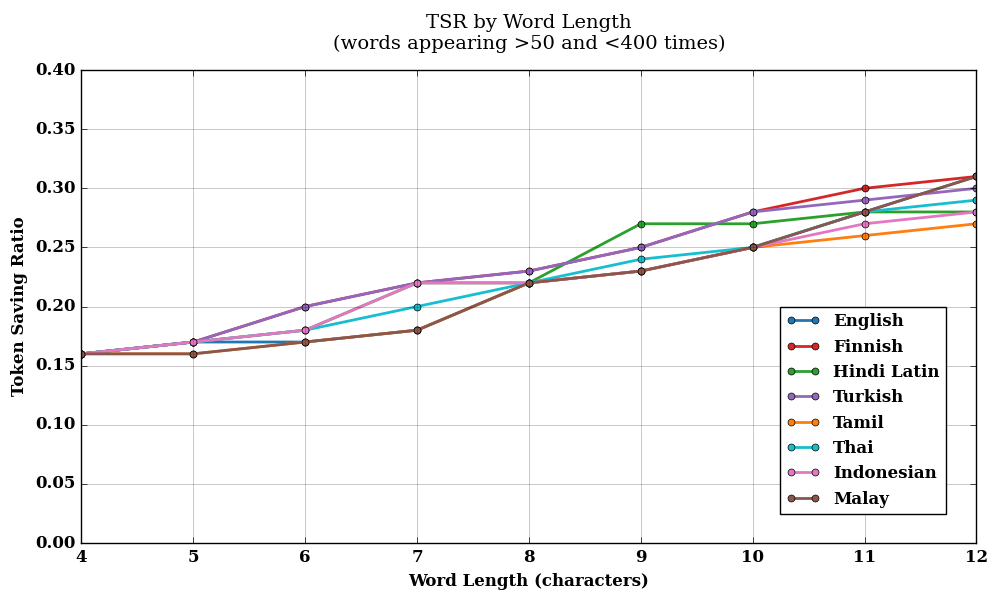}
    \caption{TSR and Word length correlation across seven different languages, with Vocab. size $m=100K$.}
    \label{fig:relative-compression}
\end{figure}
\end{center}

\noindent \textbf{Word length Relation with TSR}: We plot an interesting observation that word length has with TSR in Figure \ref{fig:relative-compression}. We notice a strong correlation, with longer words achieving better compression ratios (increasing from $\sim$0.15 for 4-character words to $\sim$0.30 for 11-12 character words) - suggesting that word length appears to be one of the factors in compression efficiency across these linguistically diverse languages. This pattern is consistent across all languages in our study, though with varying slopes - Finnish and Turkish show steeper increases with word length, while English demonstrates a more gradual rise. Notably, agglutinative languages like Finnish, Turkish, and Indonesian, which typically have longer words due to their morphological structure, benefit more from our optimal segmentation strategy as word length increases.  Thai shows a moderate but steady increase despite its analytic nature and lack of explicit word boundaries, and Tamil, with its complex agglutinative morphology, displays a more gradual rise similar to English, possibly due to its unique script-to-byte conversion patterns.

\subsection{Extrinsic Evaluation Tasks} 
Table \ref{combined_results} presents a systematic analysis across languages and tasks, examining how different types of tokenization errors—particularly compound word splitting, verb root identification, and morpheme boundary detection—affect downstream performance. For the Indonesian Emot task with the 120M model, optimal segmentation improves accuracy by 4.32\% (from 40.23\% to 44.55\%) in the full dataset. This improvement becomes more pronounced in the $TSR^*$ subset, reaching \textbf{5.64\%} (from 39.23\% to 44.87\%), primarily due to better handling of compound words (e.g., "memberikan" → "memberi" + "kan") and proper verb root preservation. In the 350M model, while the overall gap is smaller at 2.50\%, it still increases to \textbf{2.56\%} in the $TSR^*$ subset, showing similar error patterns but at reduced magnitudes. The WreTe task shows similar error patterns: optimal segmentation yields a 2.00\% improvement in the full dataset, expanding to \textbf{2.66\%} in $TSR^*$, with compound word splitting errors driving a significant portion of the performance difference. For Turkish (XNLI), we observe improvements of 0.56\% to \textbf{0.76\%} (120M) and 0.98\% to \textbf{0.79\%} (350M), where analysis shows that agglutinative morpheme boundaries (particularly case markers and possessive suffixes) significantly impact performance. Finnish presents a unique case where accuracies remain identical between \textit{All} and $TSR^*$ subsets, as all words exhibit non-zero TSR scores.

\noindent For English tasks, we observe moderate differences in the performance between \textit{All} and the $TSR^*$ subset. In Story Cloze, the 350M model shows an improvement with Optimal segmentation in $TSR^*$ (\textbf{7.83}\% gain, from 52.17\% to 60.00\%) compared to the full dataset (\textbf{0.43\%} gain, from 51.31\% to 51.74\%). QQP shows varying patterns: in the 120M model, Greedy performs better in both sets, with the gap being more pronounced in $TSR^*$ (-0.70\% vs -0.38\%). These results suggest that evaluating the $TSR^*$ subset often amplifies the impact of the segmentation strategy, particularly for tasks where token sequencing plays a crucial role.  The better performance of Greedy might be attributed to English's relatively straightforward morphological structure compared to agglutinative languages like Turkish or Finnish. English words typically have clearer boundaries and less complex internal structure, allowing the tokenization strategies to focus on semantic units rather than navigating complex morphological combinations. 

\begin{table}[!htbp]
    \centering
    \small
    \begin{tabular}{ccccccccc}
        \toprule
        \textbf{Model Size} & \textbf{Segmentation} & \textbf{Perplexity} $(\downarrow)$ \\
        \midrule
        \multirow{2}{*}{120M} & Greedy & {43.76} \\
                             & Optimal & \textbf{39.97} \\
        \midrule
        \multirow{2}{*}{350M} & Greedy & {34.56} \\
                             & Optimal & \textbf{34.45} \\    
        \bottomrule
    \end{tabular}
    \caption{GPT-2 Perplexity on English datasets (lower is better)}    
    \label{extrinsic_perplexity}
\end{table}

\noindent We also report the perplexity metric evaluation on English datasets (LAMBADA) to show that our Optimal segmentation does not substantially degrade model performance compared to Greedy segmentation. We present this result in the Table \ref{extrinsic_perplexity}. We report that the differences in perplexity are minimal. These results suggest that our proposed tokenization strategy maintains comparable modeling capability on English text, indicating that the improvements we observe on non-English tasks are not achieved at the expense of English language modeling quality.

\section{Conclusion}
\noindent In the scope of this work, we identified the inefficient greedy segmentation method used in the BPE tokenizer and proposed an optimal segmentation algorithm that results in efficient token utilization, particularly for LR languages. We established the optimality of our algorithm by showing its impact in both intrinsic and extrinsic experiments as done in the literature. By studying multiple languages, we observed a strong correlation between improvements in Token Saving Ratios and linguistically better segments, with this effect being especially pronounced for morphologically complex words and propagating to performance improvement in downstream tasks. 
These findings underscore the need for research in tokenization approaches that can boost model effectiveness, especially for language models serving low-resource languages.

\section{Limitations and Future Work}
Our work demonstrates the impact of using BPE tokenization with optimized segmentation on tokenization efficiency across multiple languages. Although we evaluated models on intrinsic metrics for a variety of languages, our extrinsic evaluations focused primarily on four languages: English, Finnish, Indonesian, and Turkish. We chose these languages to capture diversity in typology and morphology, as well as to leverage the relatively richer resources available for them compared to many other LR languages. In the future, we intend to perform a more comprehensive follow-up study to replicate these findings across a wider array of languages provided in Table \ref{tab:token_saving}, aiming to validate the broader applicability of our approach. This could help assess the robustness of using optimal segmentation across languages with more complex or less studied morphological characteristics.  

\noindent Future research would also explore other underlying factors influencing tokenization quality and its broader impact on language model success. This extension would help us understand whether our findings about optimal segmentation scale to models with larger vocabularies and more sophisticated architectures. In future work, we plan to extend our analysis to larger foundation models like LLaMA-3 \cite{grattafiori2024llama3herdmodels}, where the impact of tokenization strategies may reveal additional insights about segmentation in more complex architectures. We would also explore improvements in other stages, such as optimal vocabulary selection and encoding methods for adaptive tokenization.

\bibliography{coling}
\clearpage
\appendix

\section{Language Model Parameters}
The $120M$ parameter models were trained using the GPT architecture with the following parameters.
\begin{table}[h]
    \centering
    \small 
    \begin{tabular}{c c c c c c}
        \hline
        \textbf{Model} & \textbf{Dim.} & \textbf{Heads} & \textbf{Layers} & \textbf{Batch} & \textbf{Seq Len} \\ \hline
        120M  & 1024 & 16 & 24 & 1024 & 1024 \\ 
        350M  & 2048 & 8  & 16 & 2048 & 1024 \\ \hline
    \end{tabular}
    \caption{Model Configurations}
    \label{tab:model_config}
\end{table}

\label{sec:appendix_model}



\section{Proof of Optimality}
\label{sec:dp_proof}
\subsection{Dynamic Programming Formulation}
Define $dp[i]$ as the minimal number of tokens needed to segment the prefix $S_0 S_1 \ldots S_i$ (positions $0$ to $i$, inclusive). We set $dp[-1] = 0$ as the base case, representing the empty string requiring zero tokens.
The recurrence relation is:

\begin{align*}
dp[i] &= \min_{(0 \leq j \leq i)} (dp[j-1] + 1) 
\\ & \text{where } S_j S_{j+1} \ldots S_i \in V
\end{align*}

\subsection{Proof by Contradiction:}

Suppose there exists a segmentation of the prefix $S_0 S_1 \ldots S_i$ into tokens from vocabulary $V$ that uses fewer tokens than $dp[i]$ computed by our algorithm.

Let this supposed optimal segmentation divide the prefix into tokens, ending at positions $-1 = k_{-1} < k_0 < k_1 < k_2 < \ldots < k_{m-1} = i$, resulting in $m$ tokens:

\[
\begin{aligned}
T_0 &= S_{k_{-1}+1} S_{k_{-1}+2} \ldots S_{k_0}, \\
T_1 &= S_{k_0+1} S_{k_0+2} \ldots S_{k_2}, \\
&\ \ \vdots \\
T_{m-1} &= S_{k_{m-2}+1} S_{k_{m-2}+2} \ldots S_{k_{m-1}}.
\end{aligned}
\]

Each $T_j \in V$, and the total number of tokens is $m < dp[i]$.

Consider the last token $T_{m-1}$ in this segmentation, which covers the substring $S_{k_{m-2}+1} S_{k_{m-2}+2} \ldots S_{k_{m-1}}$. Since $T_{m-1} \in V$, our algorithm, when computing $dp[i]$, examines this possibility.

By the definition of our algorithm:

\begin{align*}
dp[i] &= \min\left(dp[i],\ dp[k_{m-2}] + 1\right)
\end{align*}

In the worst case, there are no better alternatives than $k_{m-2}$,

\begin{align*}
dp[i] &= dp[k_{m-2}] + 1
\end{align*}

By a similar argument,
\begin{align*}
dp[k_{m-2}] &= dp[k_{m-3}] + 1, \\
dp[k_{m-3}] &= dp[k_{m-4}] + 1, \\
& \vdotswithin{=} \\
dp[k_{i}] &= dp[k_{i-1}] + 1, \\
& \vdotswithin{=} \\
dp[k_0] &= dp[k_{-1}] + 1, \\
\end{align*}

Using the above results,
\begin{align*}
dp[i] &= dp[k_{m-2}] + 1 \\ 
&= dp[k_{m-3}] + 1 + 1 \\ 
& \vdotswithin{=} \\
&= dp[k_i] + m - 1 - i \\ 
& \vdotswithin{=}  \\
& = dp[k_{-1}] + m - 1 - (-1)  \\
& = m  
\end{align*}

Simplifying to,
\[
dp[i] = m
\]
However, we initially assumed that $m < dp[i]$. This leads to a contradiction, which means our initial assumption that there exists a better segmentation is wrong. This completes the proof.

\section{Intrinsic Statistical Analysis}

\begin{center}
\begin{figure}[!htbp]
    \centering
    \includegraphics[width=0.8\linewidth]{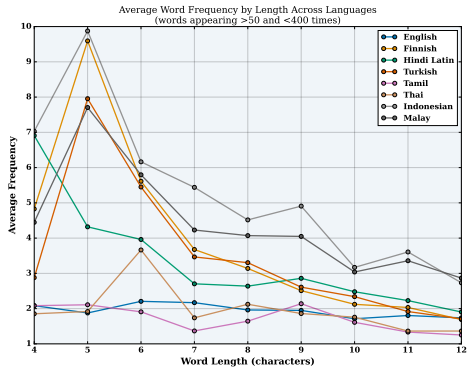}
    \caption{Frequency vs Word Length: Comparison across seven languages with a vocab size of $m=100K$
    }
    \label{fig:frequency-word}
\end{figure}
\end{center}

\noindent \textbf{Frequency analysis with Word length}: The word frequency distribution pattern provides crucial context for interpreting the extrinsic task performance. The frequency-based analysis shown in Fig. \ref{fig:frequency-word} helps explain why the impact of optimal segmentation varies significantly across languages and tasks, with larger gains in languages where optimal segmentation of longer words, though less frequent, carries greater semantic importance. The reported token saving percentages (TSR) may underestimate the true potential of optimal segmentation due to frequency-based evaluation bias. Since longer words (>6 characters) occur substantially less frequently in the corpus, their improvements in segmentation quality are numerically diluted in aggregate metrics. Many of these longer words often carry crucial semantic information through compound formation and morphological processes, as evidenced in Table \ref{tab:examples}. 

\begin{center}
\begin{figure}[!htbp]
    \centering
    \includegraphics[width=0.8\linewidth]{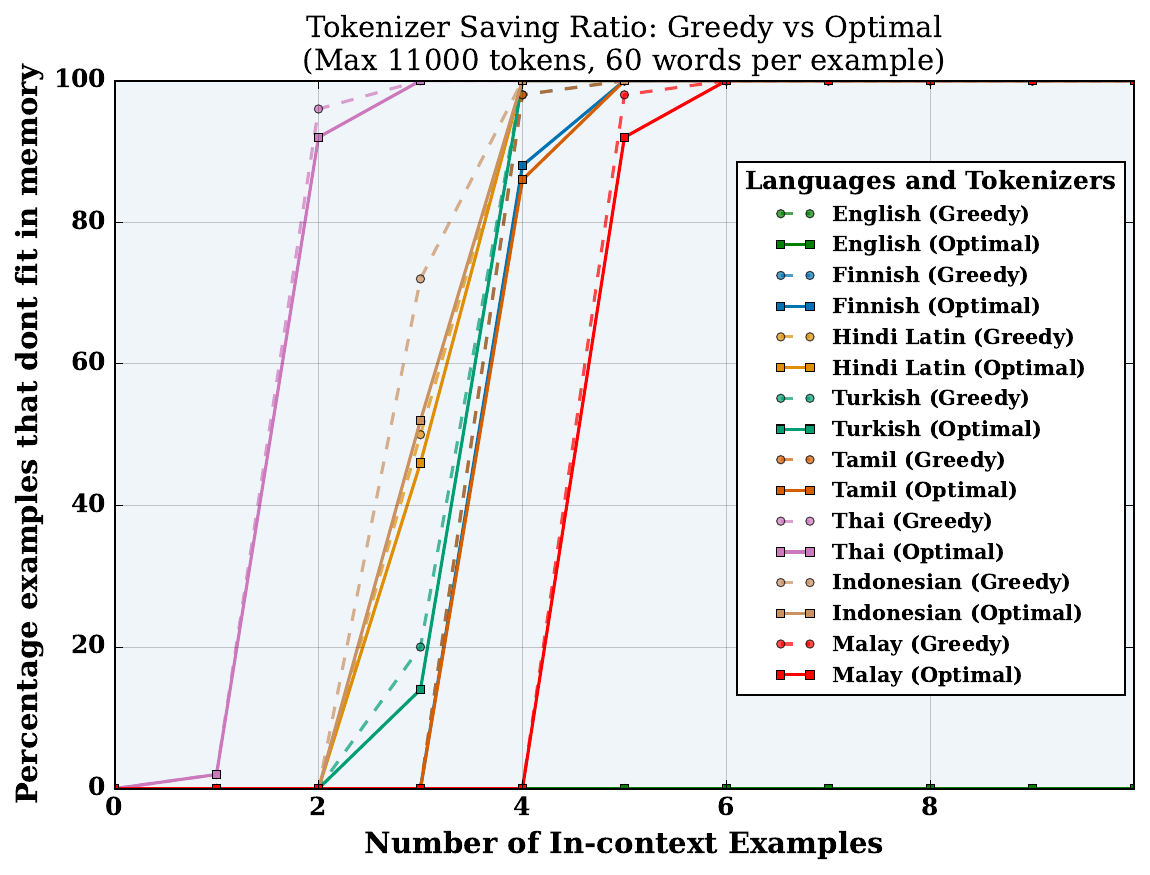}
    \caption{In-Context Comparison: Percentage of examples that fit across languages with vocab size of $m=100K$, highlighting the impact on extrinsic performance with increasing in-context examples.}
    \label{fig:tokenizer-comparison}
\end{figure}
\end{center}

\noindent \textbf{In-context evaluation:} Figure \ref{fig:tokenizer-comparison} compares the token-saving performance of greedy and optimal segmentations across different languages as the number of in-context examples increases. It shows significant variation in the token saving percentages between languages, with the Optimal tokenizer outperforming the Greedy approach. The gap between the two tends to widen as more examples are provided, indicating a better ability from a language model to leverage contextual information. This visualization offers valuable insights into the intrinsic multilingual capabilities of these tokenizers, which can inform decisions around model architecture and deployment for multilingual applications.

\section{Extrinsic Evaluation Tasks} 

\label{sec:appendix-extrinsic}
We describe the different tasks used for fine-tuning our models:
\begin{itemize}
  \item For English generation tasks, we used the Penn Tree Bank (PTB) dataset \citep{ptb}, which serves as a traditional benchmark for assessing language generation capabilities through zero-shot perplexity, leveraging its pre-internet content. Additionally, the LAMBADA dataset \citep{lambada} was employed to test the model's ability to comprehend and predict the last word in a paragraph, challenging its handling of long-range dependencies. For English classification tasks, we utilized the Quora Question Pairs (QQP) dataset \footnote{\url{https://quoradata.quora.com/First-Quora-Dataset-Release-Question-Pairs}}), which involves determining if question pairs are duplicates, evaluated using the F1 metric.  The Story Cloze dataset \citep{story_cloze} was also used to measure the model’s ability to choose the correct ending for short narratives, further assessing classification performance.
  \item For Finnish we used the gold passage version of the Typologically Diverse Question Answering dataset (\textbf{TyDiQA-GoldP}) \cite{tydiqa} \cite{ruder-etal-2021-xtreme}. It consists of a question, a relevant passage, and an answer - yes or no.
  \item Expanding to Indonesian, we employed two datasets from the indoNLU \citep{indonlu, emot, wrete} collection : EmoT, which is an emotion classification dataset collected from Twitter consisting of tweets in Indonesian covering five emotion labels: anger, fear, happiness, love, and sadness; and WReTE, which is a textual entailment dataset constructed from Wikipedia revision history, containing pairs of sentences with binary semantic relations .
  \item For Turkish, the XNLI dataset \citep{xnli} was utilized. XNLI extends the MultiNLI dataset into a multilingual evaluation suite, providing a benchmark for cross-lingual language understanding through sentence-pair classification tasks across 15 languages.
\end{itemize}

\end{document}